# Jokes Aside: Measuring the Semantic Distance of Double Meanings

Fabio De Ponte

**Université de Namur**
fabio.deponte@unamur.be
*ORCID 0000-0001-7139-3057*

## ABSTRACT

*Large language models have significantly enriched the toolkit for computational humor research, particularly in the automated generation of jokes and puns. A key innovation, contextual embedding vectors, offers new opportunities to revisit and refine earlier hypotheses. Notably, Petrovic and Matthews (2013) proposed a joke generation model based on the scheme "I like my X like I like my Y, Z" (e.g. "I like my ice like I like my dreams, crushed"). They suggested that joke hilarity increases with: a) frequent association of Z with X and Y, b) rarity of Z, c) ambiguity of Z, and d) meaning distance between X and Y. Building on this, Winters et al. (2019) proposed a set of metrics, based on Google Ngrams and Word2Vector. In this work, three out of their five metrics are revisited with word embeddings: obviousness, compatibility, and comparison. Another measure, symmetry, defined as closeness of Z to both X and Y, is introduced here for the first time. Two models were used to collect the embedding vectors (OpenAI text-embedding-3-small and MiniLM all-MiniLM-L6-v2) on three datasets: JokeJudger, Expunations, and rJokes. The last two datasets, Expunations, and rJokes, were expanded by adding paired sentences that captured the ambiguous expression at the core of each joke in its two different meanings. Results revealed that models trained on the proposed metrics performed poorly in predicting humor ratings: on JokeJudger, the best model achieved 57.1% accuracy, below the 61.5% baseline, while performance on Expunations and rJokes was even lower. Nevertheless, the symmetry metric seems consistently associated with higher-rated jokes, suggesting it may capture a necessary—though not sufficient—property of humor.*

## INTRODUCTION

Humor is a distinctive trait of human communication and with the rise of the Internet, it has come to permeate nearly every aspect of online life. From casual exchanges on social media to political satire, humor shapes interactions and strengthens social bonds. Humor spreads quickly and widely in the form of memes, irony, and viral jokes, often capturing cultural and societal dynamics in real time. Yet despite its ubiquity, automated humor recognition remains a challenge. In recent years, advances in Natural Language Processing (NLP) – particularly the development of large language models (LLMs) – have renewed interest in the topic, offering the promise of overcoming longstanding obstacles. But one of the main challenges in studying humor is that people rarely agree on what is funny, making it difficult to obtain reliable data for modeling. Humor is inherently subjective and constantly evolving: it's a moving target.

## BACKGROUND

The study of humor dates back to ancient Greece, with the **superiority theory**—attributed to Plato and later developed by Hobbes and Descartes (Morreall, 2024) —which posits that laughter arises from a sense of superiority over others or over a former version of ourselves. In the eighteenth century, an alternative view was proposed by Lord Shaftesbury in his 1709 *Essay on the Freedom of Wit and Humor*, where he suggested that humor expresses "the natural free spirits of ingenious men," which, if repressed, will "find out other ways of motion to relieve themselves in their constraint." This idea evolved into the **relief theory**, later embraced and expanded by Spencer (1875), who described laughter as the physical release of repressed emotions in the form of "nervous energy." Freud (1905 [1960]) further developed this perspective, focusing on the connection between jokes and the unconscious mind.

During the same period, a more cognitively oriented explanation emerged: the **incongruity theory**, which argues that humor arises from the apparent contradiction between two elements, resolved by a third—typically a punchline—that reinterprets the initial conflict. This theory was supported by Kant (1790), Schopenhauer (1844), and Kierkegaard (1846). Later theorists, such as Eysenck (1942) and Suls (1972), built on these ideas by suggesting that jokes often follow a two- or three-stage structure culminating in a punchline.

**The "I like my X like I like my Y, Z" joke template**

In recent years, many attempts have been made at supporting theories with quantitative measures and models. One of them came from Petrovic and Matthews (2013), who proposed a model to generate jokes in the form "I like my X like I like my Y, Z" (e.g. *"I like my ice like I like my dreams, crushed"*) where X are Y were often two nouns and Z an adjective. The model, based on the incongruity theory, leveraged the frequency of co-occurrence of words in large corpus and was based on the assumption that the degree of hilarity of a joke increases with:

i. The increase of frequency of the use of the attribute contained in Z (*"crushed"*) to describe both X (*"ice"*) and Y (*"dreams"*).
ii. The decrease of the frequency of the general use of Z (*"crushed"*); in other words, its rarity.
iii. The increase of the degree of ambiguity of the attribute.
iv. The increase of the distance between X (*"ice"*) and Y (*"dreams"*).

To measure word co-occurrence frequencies, the authors used Google Ngrams, while they turned to WordNet to assess the number of possible interpretations—and thus the degree of ambiguity—of specific words.

Of the jokes generated by the model, 16.3% were considered funny by human raters. While this might seem modest, it is noteworthy in comparison to the 33% of jokes produced by humans using the same template that were considered funny by the same raters.

Building on that work, Winters *et al.* (2019) introduced **Goofer**, a model that, given a noun X, generated a compatible adjective Z, then produced another noun Y that could be associated with it. A rater model was then applied, and only jokes rated above a certain threshold were published. The final result was that 11.4% of the generated jokes were considered funny. While this rate was lower than that of Petrovic and Matthews' model, Winters and colleagues noted that the rating scales differed: the first model used a 3-point scale, while the second employed a more fine-grained 5-point scale. Under this new classification system, the percentage of human-generated jokes rated as funny also decreased, dropping to 27.4%.

More importantly, as mentioned, to select the jokes to be published, the team developed a rating model based on **Incongruity-Resolution theory**, as proposed by Ritchie (1999), which identifies five key elements in the relationship between the first and second interpretation:

- **Obviousness:** Refers to how natural and straightforward the initial interpretation appears.
- **Conflict:** Arises when the punchline contradicts the initial meaning, triggering the need for reinterpretation.
- **Compatibility:** Ensures the punchline aligns with a second, hidden meaning.
- **Comparison:** Emphasizes the necessity for distinct interpretations.
- **Inappropriateness:** Suggests that the second meaning should feel odd, inappropriate, or taboo, thereby enhancing the humorous effect.

Winters *et al.* (2019) proposed five metrics to evaluate these elements, relying on Google Ngrams and WordNet, and for the latter, also incorporating the DEViaNT system (Kiddon and Brun, 2011):

- **Word Frequency**: Derived from Google Ngrams' 1-gram data, this metric was used to estimate the **Obviousness** of a word.
- **Word Combination Frequency (2-grams)**: Focusing on adjective-noun pairs, this metric was used to assess both **Conflict** and **Compatibility**. The pairs were filtered using WordNet and a POS tagger to retain only relevant combinations.

- **Number of Meanings**: Based on the number of definitions available for a word in WordNet, this metric related to **Compatibility**.
- **Adjective Vector Similarity**: This metric measured how similarly adjectives are distributed across nouns and supports the **Comparison** property.
- **Word Sexiness**: Calculated by comparing word frequency in a sexual corpus (from textfiles.com) to its frequency in general usage, this metric estimated the **Inappropriateness** of a word.

Several studies related to various forms of incongruity theory have utilized **Word2Vec**. Yang *et al.* (2015) applied it to measure semantic disconnection within sentences, while Morales and Zhai (2017) leveraged it to incorporate background text sources. Ziser *et al.* (2020) adopted a similar approach to highlight feature differences between questions. Additional studies, such as those by Cao (2021) and Huang *et al.* (2022), have also employed Word2Vec for humor analysis.

**From Ngrams to embeddings**

In a separate line of research, the advent of **word embedding vectors**—which capture certain relational aspects between words as they are used by a community of speakers—has opened new areas for investigations.

One of the earliest works with Word Embeddings (WEs) and humor, by Engelthaler and Hills (2017), focused on single words. Through a study on five thousand nouns, they found that some words are consistently considered more humorous than others. Gultchin *et al.* (2019) expanded on their work and suggested that a person's sense of humor can be represented as a vector by averaging the embeddings of the words they find funny. This personalized humor vector could then be used to predict which new words someone is likely to find humorous. Moreover, the authors clustered vectors of multiple demographic groups, and found that different humor preferences emerge across different groups.

More recently, many works, relying on word embeddings, focused on detecting specific kinds of humor, such as sarcasm (Ghosh, Guo and Muresan, 2015; Ghosh and Veale, 2016; Eke, Norman and Shuib, 2021; Ahuja and Sharma, 2022; Misra and Arora, 2023), sometimes with multimodal approaches (Chauhan *et al.*, 2022; Bedi *et al.*, 2021; Sharma *et al.*, 2020), or punchline detection (Choube and Soleymani, 2020) or self-deprecating humor (Kamal and Abulaish, 2020). For a systematic review of recent computational approaches for humor classification, see Kenneth (2024).

Tasnia *et al.* (2023) introduced a neural network architecture that improves textual context representation by applying a stacked embeddings strategy above an LSTM layer, enabling the identification of humorous and ironic content in text. A similar approach was used by Annamoradnejad and Zoghi (2022), who divided jokes into text chunks and generated embeddings for each chunk using the BERT model. These embeddings were processed through parallel hidden layers in a neural network—one stream per sentence—to extract latent features. The outputs were then concatenated to capture inter-sentence congruity and other relational patterns, ultimately predicting humor ratings. The model achieved F1 scores of 0.982 and 0.869 in humor detection tasks, demonstrating that transformers can model the latent spaces and relational patterns that enable humor recognition.

In fact, even a simple interaction with ChatGPT suggests that models like OpenAI's GPT-4 can manage humor. For instance, when prompted with, *"Explain: What do you call it when Batman skips church? Christian Bale!"* the model responds: *"It is a pun that plays on the actor's name, Christian Bale, who played Batman. It combines the idea of a 'Christian' (someone who might go to church) who 'bails' (slang for skipping), turning that phrase into the actor's name. The humor comes from the double meaning and sound similarity."*

However, the models lack interpretability. Their use does not help confirm or refute theories about the underlying mechanisms of humor. This is the problem addressed by the work presented here: rather than using embedding vectors as part of a larger model that predicts ratings, they are used to calculate specific metrics. This approach allows to investigate the validity of humor theories, offering deeper insights into the mechanisms behind humor, rather than merely predicting results.

## METHODOLOGY

The model presented here[1] is inspired by the five elements suggested by Ritchie (1999) and transformed in metrics by Winters *et al.* (2019). In particular, with reference to the template "I like my X like I like my Y, Z," (for example, in *"I like my sex like I like my police officers, with body cams"*) four metrics are proposed:

- **Obviousness**: the cosine similarity of the embedding vectors of X (*"sex"*) and Z (*"with body cams"*).

For the pun to work, *this should be high*.

- **Compatibility**: the cosine similarity of the embedding vectors of Y (*"police officers"* and Z (*"with body cams"*).

For the pun to work, this *should be high*.

- **Comparison**: the cosine similarity of the embedding vectors of XZ (*"sex with body cams"*) and YZ (police officers with body cams").

For the pun to work, this *should be low*.

- **Symmetry**: the difference between obviousness and compatibility. This metric replaced conflict.

For the pun to work, this *should be low*.

The last metric addresses the question: do the two meanings work equally well? Consider the example *"Why did the photon check a bag at the airport? It didn't—it was traveling light!"* The pun relies on the double meaning of the word "light," which can be both a noun (representing the physical phenomenon that enables vision) and an adjective (meaning the opposite of heavy). The humor comes from the fact that the expression "traveling light" can be easily interpreted in both ways—its double meaning is symmetrical.

In contrast, consider this example: *"Would you like some soda in your whiskey? asked Tom caustically."* The play here is on "caustic soda," but the problem is that "caustically" is not a common adverb used in reference to soda. To understand one side of the pun, a person must force their mind to connect two different forms to a single meaning, disrupting the symmetry and making the double meaning less effective.

Two **models** in particular were used to collect the embedding vectors: **OpenAI text-embedding-3-small**[2] and **MiniLM all-MiniLM-L6-v2**[3].

1 Source code and data are available at: github.com/fabiodeponte/JokesAside.

2 Available at: platform.openai.com/docs/models/text-embedding-3-small.

3 Available at: huggingface.co/sentence-transformers/all-MiniLM-L6-v2.

In addition to **JokeJudger**[4], the dataset introduced by Winters *et al.* (2019), the four metrics were applied to two other datasets: **Expunations**[5] (Sun *et al.,* 2022) and **r/Jokes**[6] (Weller and Seppi, 2020).

### The JokeJudger dataset

Introduced by Winters *et al.* (2019), the JokeJudger dataset contains 521 jokes, including 100 generated by Goofer, a model developed by the authors to create jokes in the format "I like my X like I like my Y, Z." Each joke is represented by three texts (X, Y, and Z) and is accompanied by a vector containing ratings. The ratings were collected through an online platform created by the authors for the experiment. The individual ratings range from 1 to 5, with each joke rated by a variable number of people, ranging from 4 to 57. Training a random forest with their metrics, Winters *et al.* (2019) were able to reach an accuracy on rating prediction of 61.5%, which we will consider our baseline.

For the purposes of this work, in order to calculate the mentioned four metrics, four steps were taken:

1) The rating was averaged and approximated to a class of 1, 2, 3, 4, or 5 stars.
2) Two sets of embedding vectors, one with MiniLM all-MiniLM-L6-v2 model and one with OpenAI text-embedding-3-small model were calculated for X, Y, Z, XZ, and YZ.
3) Obviousness, compatibility, symmetry and comparison metrics were calculated.
4) With those four metrics and the target rating calculated at step 1, several models were trained: random forest, support vector machine, naive Bayes, and regression.

The results are shown in Table 1. The highest performance was achieved by a random forest model, with parameters defined through a 5-fold cross-validation training process, which resulted in a test accuracy of 57.1%.

| Model | Embedding Vectors | Results |
|---|---|---|
| Random forest | MiniLM | Accuracy: 57.1% |
| Regression | MiniLM | $R^2$: -0.02 |
| SVM | MiniLM | Accuracy: 48.6% |
| Naive Bayes | MiniLM | Accuracy: 53.3% |
| Random forest | OpenAI | Accuracy: 54.3% |
| Regression | OpenAI | $R^2$: -0.04 |
| SVM | OpenAI | Accuracy: 51.4% |
| Naive Bayes | OpenAI | Accuracy: 52.4% |

*Table 1: Performances of the models on JokesJudger dataset*

4 Available at: github.com/twinters/jokejudger-data/tree/master.

5 Available at: github.com/amazon-science/expunations.

6 Available at: github.com/orionw/rJokesData.

### The Expunations dataset

Expunations was introduced by Sun *et al.* (2022) at the Conference on Empirical Methods in Natural Language Processing (EMNLP) in 2022. It is an annotated selection of puns from the SemEval 2017 Task 7 dataset (Miller *et al.*, 2017). The authors enhanced the dataset with crowdsourced annotations, including keywords that highlight the most distinctive words contributing to the humor, explanations of why the text is funny, and fine-grained funniness ratings. The dataset contains 1,999 puns, each annotated with 5 funniness ratings (ranging from 1 to 5). Raters were also invited to provide natural language explanations, so most puns in the dataset include at least one explanation. Additionally, it includes answers to the question, "Do you understand the text?"

Here's an example: *"When the glazier was sent to the hospital room to check the cracked window, he told the patient in the body cast, 'I've come to feel your pane.'"* Each pun is accompanied by one or more explanations, such as: *"This is a play on words. 'Pane' is a sheet of glass that a glazier would fit to a window. However, 'pain' refers to physical discomfort often felt by a hospital patient."*

Finally, each pun is tagged with relevant keywords, such as: "glazier," "window," "feel your pane," "patient in the body cast," "hospital," "cracked window," and others.

In the course of the work presented here, the dataset was further expanded with four additional fields: the expression at the center of the pun in two versions (to cover the cases of homophones), and two sentences that use the expression in both of its implied meanings. For example, in the case of the pun above, the expression "pane" and "pain" were included. Two sentences were added: one for each meaning of the expression. The sentences were: "The glazier installs a pane in the window." (for "pane") and "He was in a lot of pain while he was at the hospital." (for "pain").

The content was generated by iteratively querying the OpenAI model (o3-mini-2025-01-31) with the pun, its explanations, keywords, and the following prompt:

> *Below, you can find a pun, based on the double meaning of an expression, along with it a unique identifier and, if available, some explanations and some keywords.*
>
> *I want you to:*
>
> *- select the expression at the core of the joke. If it is a homophone, select the two versions of it.*
>
> *- create two serious (without humor) sentences with the same expression: one where it takes the first meaning, and one where it takes the second meaning.*
>
> *The result of the computation must be returned in a format like this, to fit a CSV file separated by a semicolon:*
>
> *ID; expression1; expression2; sentence1; sentence2.*
>
> *Note: if the expression is a homophone, write the two versions. Otherwise, just write the same expression twice.*

The result was a dataset of 1,895 puns with the above-mentioned fields. Subsequently, two sets of embedding vectors (for the pun, the two forms of the central expression, and the two sentences) were calculated: one using the aforementioned MiniLM model and the other using the OpenAI model.

Using these embeddings, the four metrics were calculated according to the following scheme:

- **Obviousness**: the cosine similarity of the embedding vectors of expression 1 ("pane") sentence 1 (*"The glazier installs a pane in the window."*).
- **Compatibility**: the cosine similarity of the embedding vectors of expression 2 ("pain") and sentence 2 (*"He was in a lot of pain while he was at the hospital."*).
- **Symmetry**: the difference between obviousness and compatibility.
- **Comparison**: the difference between two cosine similarities, on one side the cosine similarity between the text of the pun (*"When the glazier was sent to the hospital room to check the cracked window, he told the patient in the body cast, I've come to feel your pane"*) and the sentence 1 (*"The glazier installs a pane in the window."*); and on the other side, the cosine similarity between the text of the pun and the sentence 2 (*"He was in a lot of pain while he was at the hospital."*).

The comparison metric is slightly different from the one adopted with the JokesJudger dataset, due to the differences in the dataset. Its aim, however, remains to measure the distance between the two interpretations of the pun.

The table 2 below shows the results.

| Model | Embedding Vectors | Results |
|---|---|---|
| Random forest | MiniLM | Accuracy: 49.7% |
| Regression | MiniLM | $R^2$: -0.01 |
| SVM | MiniLM | Accuracy: 51.1% |
| Naive Bayes | MiniLM | Accuracy: 48.9% |
| Random forest | OpenAI | Accuracy: 50.7% |
| Regression | OpenAI | $R^2$: -0.01 |
| SVM | OpenAI | Accuracy: 51.0% |
| Naive Bayes | OpenAI | Accuracy: 50.3% |

*Table 2: Performances of the models on Expunations dataset*

In this case, the best results were achieved through a SVM, that got a test accuracy of 51.1%.

### The rJokes dataset

The rJokes dataset, introduced by Weller and Seppi (2020), contains over 550,000 jokes shared on the forum Reddit r/Jokes subreddit over an 11-year span. It includes quantitative indicators of humor based on user feedback from the community. Jokes that are found funny tend to receive upvotes, while less appreciated ones are downvoted. According to the authors, while this system doesn't perfectly capture the quality of humor, it offers a useful approximation: a joke with only a few upvotes is probably less amusing to most users than one with 10,000.

The ratings range from 1 to several thousand. For the purposes of this work, they were mapped into five classes: 0 (class 0), 1–10 (class 1), 11–100 (class 2), 101–1,000 (class 3), and over 1,000 (class 4).

A selection of jokes based on word double meanings was extracted from the dataset. After a preliminary division of the dataset into several CSV files containing 100 jokes each, the selection was made through an extensive series of requests to the OpenAI model (o3-mini-2025-01-31), using the following prompt:

*In the csv file, you can find a list of jokes. I want you to select only the ones where the joke is based on a word that has these characteristics:*

*- In the context of the joke, the word takes two different meanings.*

*- It is spelled exactly the same way in both meanings: discard jokes base on homophones.*

*- It's a one single word - no composite words.*

*- It's a word present in the dictionary.*

*For each sentence, then I need you to create two serious (without humor) sentences with the word at the center of the joke: one where the word takes the first meaning, and one where it takes the second meaning.*

*The result of the computation must be returned in a format like this, to fit a CSV file separated by a semicolon:*

*sentence; word; explanation; first new sentence; second new sentence*

This process resulted in 479 jokes, each expanded with the central expression and two sentences—one for each meaning of the expression.

The jokes were further filtered based on their structure: only those consisting of two parts, a setup and a punchline, separated by a question mark, were selected and split into two separate fields. The final dataset comprised 263 jokes, formatted as the example that follows.

**Setup:** *"What did the debater say after getting stabbed during an argument?"*

**Punchline:** "good point"

**Central expression:** "point"

**Explanation:** "A point can refer to a sharp end or tip of an object. It can also mean an argument or assertion in a discussion."

**Sentence 1:** "The knife had a sharp point that could easily cut through paper."

**Sentence 2:** "During the debate, he made a valid point that swayed the audience."

**Score:** 219.0

**Score_class:** 3

This is the distribution of the score classes:

| Class | Frequence |
|---|---|
| 0 | 63 |
| 1 | 105 |
| 2 | 81 |
| 3 | 8 |
| 4 | 6 |

*Table 3: Class frequency of rJokes dataset*

The four metrics were calculated on this dataset, with minor accommodations:

- **Obviousness**: the cosine similarity of the embedding vectors of setup ("What did the debater say after getting stabbed during an argument?") and the sentence 1 ("The knife had a sharp point that could easily cut through paper.")
- **Compatibility**: the cosine similarity of the embedding vectors of punchline ("good point") and sentence 2 ("During the debate, he made a valid point that swayed the audience.")
- **Symmetry**: the difference between obviousness and compatibility.
- **Comparison**: the cosine similarity between the text of the setup and the punchline.

Results are reported in table 4. In this case, as with the Expunations dataset, the best result (an accuracy of 41.8%) was achieved through SVM.

| Model | Embedding Vectors | Results |
|---|---|---|
| Random forest | MiniLM | Accuracy: 36.7% |
| Regression | MiniLM | $R^2$: -2.05 |
| SVM | MiniLM | Accuracy: 41.8% |
| Naive Bayes | MiniLM | Accuracy: 27.8% |
| Random forest | OpenAI | Accuracy: 36.7% |
| Regression | OpenAI | $R^2$: -1.14 |
| SVM | OpenAI | Accuracy: 41.8% |
| Naive Bayes | OpenAI | Accuracy: 38.0% |

*Table 4: Performances of the models on rJokes dataset*

## DISCUSSION

On the JokeJudger dataset, the best performance was achieved using a random forest trained on metrics derived from embedding vectors generated by the MiniLM model (*all-MiniLM-L6-v2*). The model reached an accuracy of 57.1% on a classification task where scores were discretized into five classes—falling short of the 61.5% baseline established by Winters *et al.* (2019).

Winters *et al.* used five metrics grounded in the Incongruity-Resolution theory: **obviousness** (how natural the initial interpretation is), **conflict** (the degree of contradiction between interpretations), **compatibility** (how well the punchline supports a hidden second meaning), **comparison** (the semantic distance between interpretations), and **inappropriateness** (the extent to which the second meaning is odd or taboo). These were computed using resources such as Google Ngrams and WordNet, which provided word frequency, co-occurrence statistics, and the number of word senses. The final metric—**word sexiness**—was based on word frequency within a corpus of sexual texts.

| Metric | Winters *et al.* (2019) | This work |
|---|---|---|
| Obviousness | Word Frequency with Google Ngrams 1-gram of X | Cosine similarity between the embedding vectors of X and Z |
| Compatibility | *Word combination frequency with Google Ngram 2-gram of Y and Z* | Cosine similarity between the embedding vecots of Y and Z |
| Conflict | Word combination frequency with Google Ngram 2-gram of X and Y | Replaced by Simmetry: the difference between obviousness and compatibility |
| Comparison | Measure with WordNet of the number of meanings of X and Y | Cosine similarity between the embedding vectors of the composition XZ and the composition YZ |
| Inappropriateness | Word frequency in a sexual corpus (from textfiles.com) versus general usage | Discarded |

*Table 5: Summary of the metrics*

In the present work, a similar analysis was carried out on the same dataset, using three of the original metrics—**obviousness**, **compatibility**, and **comparison**—but replacing Ngram and WordNet statistics with word embedding vectors. The **conflict** metric was replaced by **symmetry**, while **inappropriateness** was excluded.

Unexpectedly, the use of embedding-based metrics did not outperform the simpler, earlier methods. Indeed, as shown in Figure 1, the distribution of the **symmetry** metric shows no clear correlation with humor ratings—suggesting that the assumed relationship between symmetry and perceived funniness may be weaker than hypothesized, or at least not effectively captured by the embedding-based approach.

As shown in Figures 2, 3, and 4, the other three metrics also displayed no clear correlation with humor scores. A regression analysis using all four metrics to predict the average score yielded a negative $R^2$ value of -0.02 on the test dataset, indicating a complete lack of explanatory power and confirming that no meaningful relationship could be established between the metrics and the humor ratings.

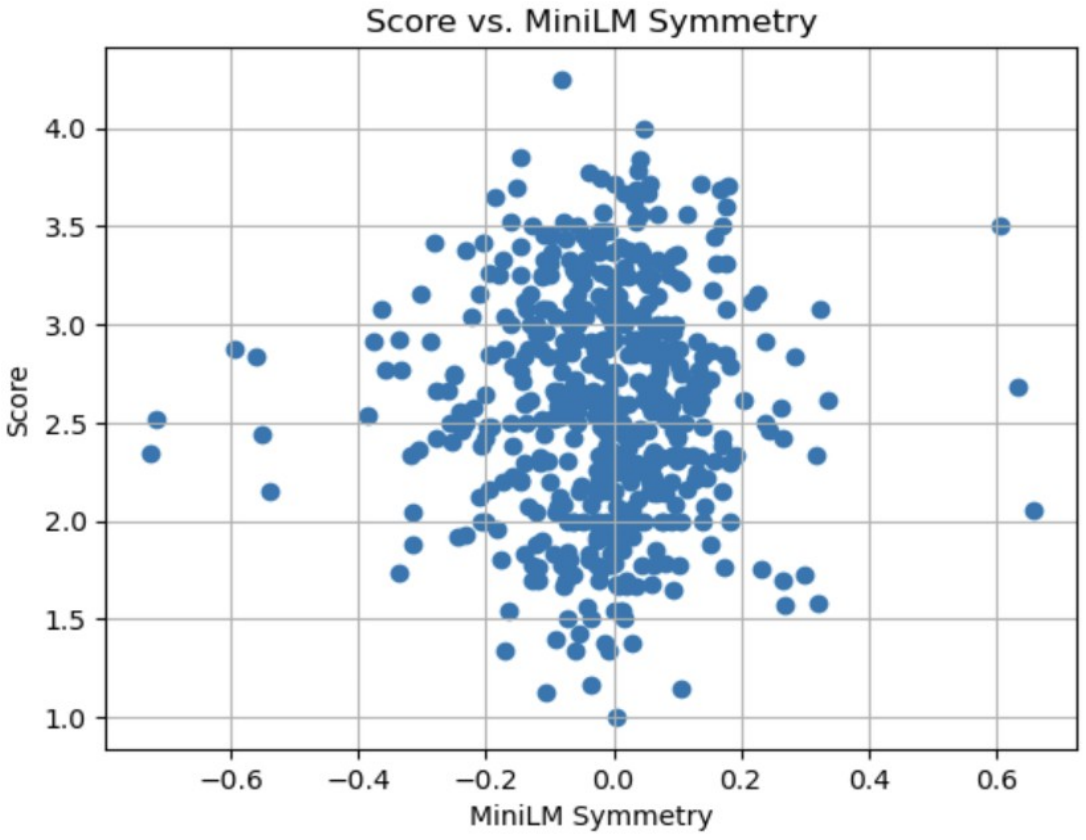

*Figure 1: Score vs. Symmetry (MiniLM model – JokeJudger dataset)*

Applying the four metrics to the other two datasets, *Expunations* and *rJokes*, did not yield improved results. In both cases, the highest classification accuracy—51.1% for *Expunations* and 41.8% for *rJokes*—was achieved using a support vector machine (SVM) with embeddings from the MiniLM model and lower than the baseline. Once again, regression analysis performed poorly, with both datasets producing negative $R^2$ values, indicating no predictive value.

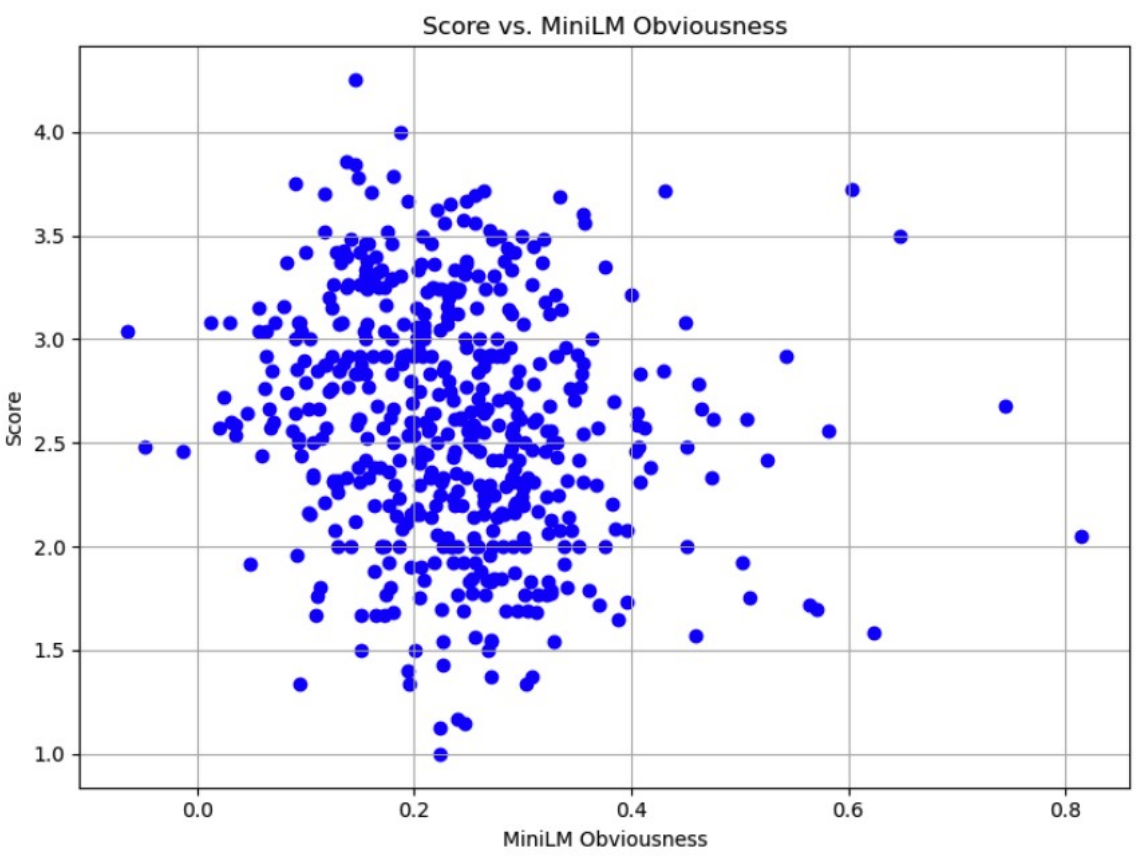

*Figure 2: Score vs. Obviousness (MiniLM model – JokeJudger dataset)*

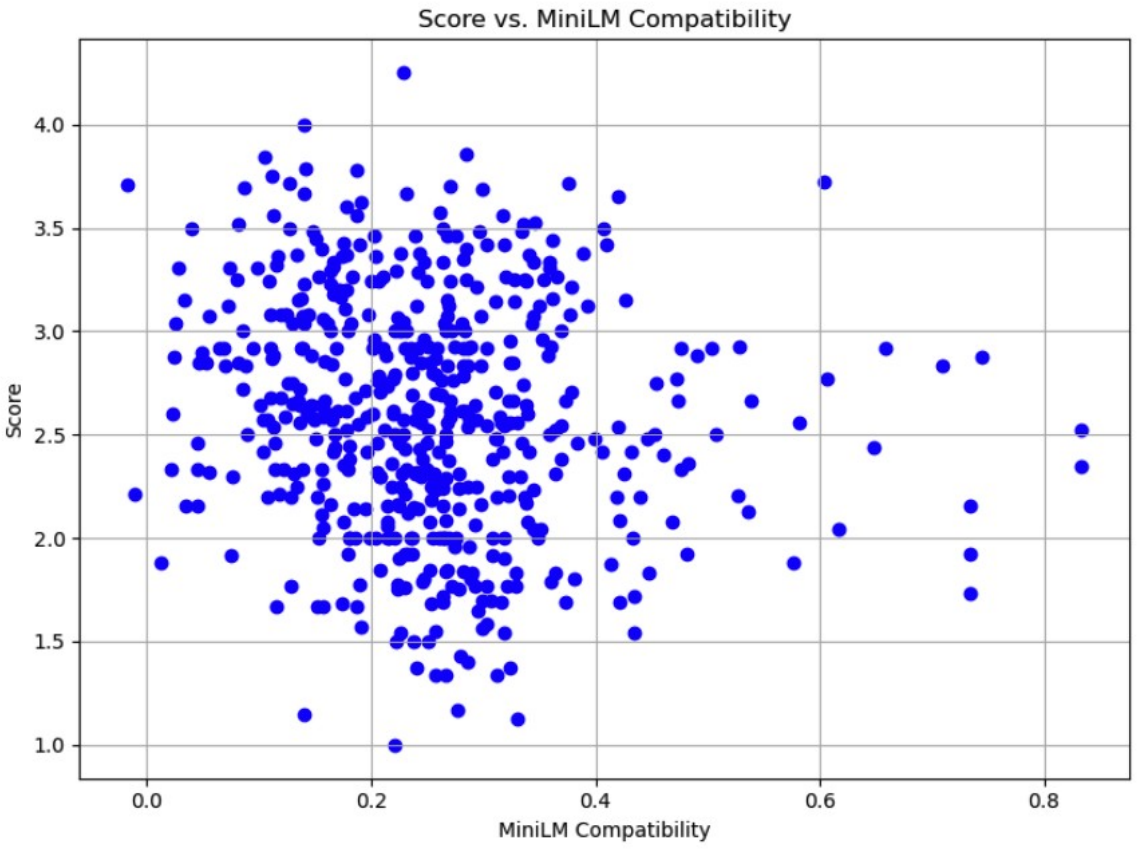

*Figure 3: Score vs. Compatibility (MiniLM model – JokeJudger dataset)*

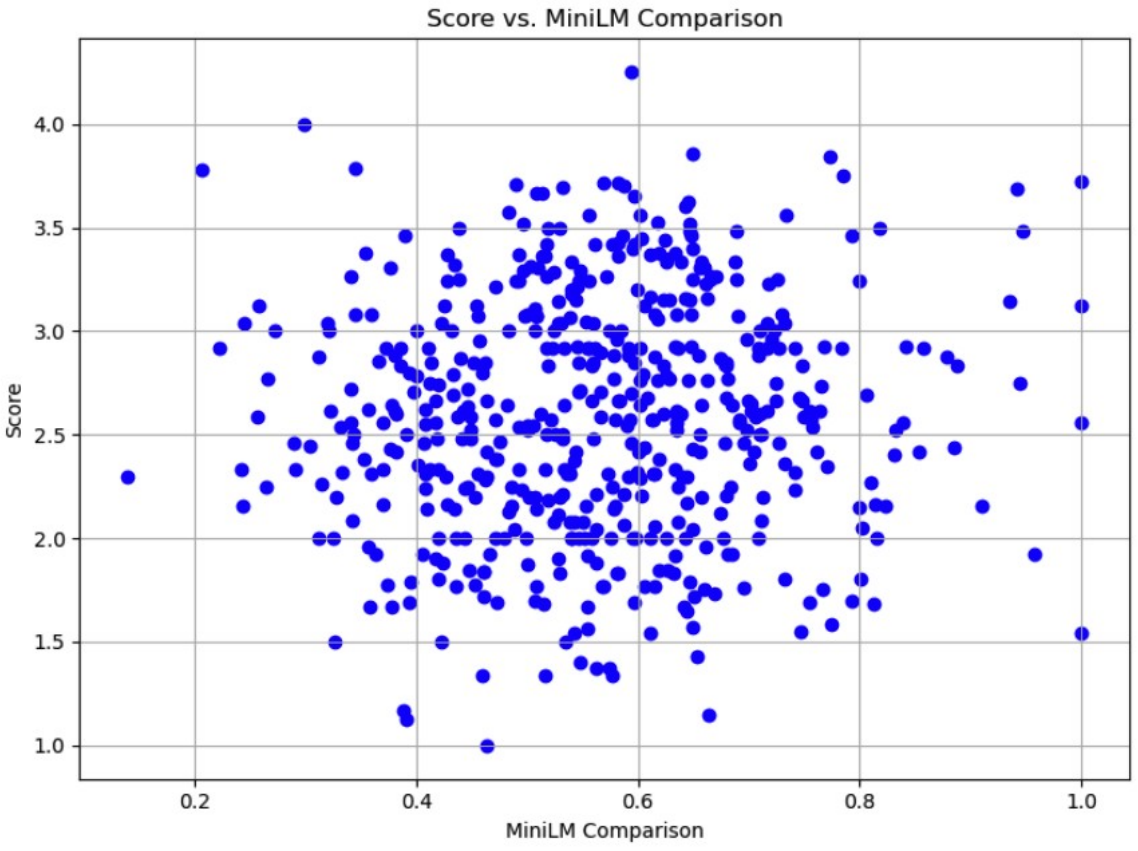

*Figure 4: Score vs. Comparison (MiniLM model – JokeJudger dataset)*

The poor performance of the explained metrics can be due to several possible reasons:

- Most of the jokes in the three datasets are based on heterographic **homophones**—expressions that sound the same but differ in both spelling and meaning. These phonologically identical yet semantically and graphically distinct terms (e.g., *"What do you call a sleeping prisoner? Under a-rest."*) lose their ambiguity in written form, since only one spelling can be represented at a time. As a result, embedding models trained on written text tend to capture just one of the two meanings, distorting any metric—such as symmetry—that relies on a balanced representation of both interpretations. To address this limitation, the expanded version of the Expunations dataset included each ambiguous expression in both of its written forms, each corresponding to one of the meanings. For example, in the pun *"Romance isn't a science, it's a heart,"* the final word was encoded both as "heart" and "art" to ensure that the semantic representations captured the full duality of the joke.

- Some jokes are **not based on wordplay** at all. For instance, in *"How do you stop a lawyer from drowning? Shoot him before he hits the water,"* there is no semantic ambiguity involving a specific word. Instead, the humor arises from an unexpected reinterpretation of the scenario, rather than from a second meaning of any term in the setup. The punchline shifts the perspective entirely, offering a stark contrast in intent rather than exploiting lexical ambiguity.

- Many jokes draw on cultural **stereotypes** or biases that are unlikely to be fully captured by word embeddings—if they are captured at all.

- Some jokes rely on references to **celebrities**, such as: *"What do you call it when Batman skips church? Christian Bale."* Christian Bale is the actor who portrayed Batman in *The Dark Knight* trilogy. The punchline plays on the homophonic phrase "Christian bale," which sounds like "a Christian who bails"—with "bail" being slang for skipping or abandoning something. The humor in this case depends on both phonetic ambiguity and shared cultural knowledge about the actor and the character he played.

- Many jokes also feature **sexual references**. For instance, as Ritchie (1999) points out, the humor in jokes like *"How did the farmer find* *the sheep in the tall grass? Very satisfying"* is heightened by its inappropriateness. However, as noted earlier, the metric of inappropriateness was not included in the present study.

- Contextual embeddings may not be directly comparable using cosine similarity as straightforwardly as this study assumes. Furthermore, even if they were, the two selected models might not be the most suitable for the task.
- Using the OpenAI model *o3-mini-2025-01-31* both to select jokes based on certain properties and to generate sentences intended to capture their dual meanings may introduce fragility into the system. It is worth noting, however, that this setup—if it has any impact—is more likely to result in overfitting, potentially inflating the performance of the metrics rather than diminishing it.
- The datasets may contain significant noise. A closer inspection reveals a heterogeneous use of punctuation, text emoticons, and acronyms like NSFW (Not Safe For Work), a warning label commonly used online to indicate that a piece of content — such as a joke, image, video, or article — contains material that may be inappropriate or offensive in professional or public settings.
- The JokeJudger dataset ratings were collected through an online platform open to anyone interested, with no random sampling of demographic groups, while rJokes ratings were extracted as the number of upvoting a joke collected on the social network Reddit. In both cases, the average rating may not be representative of the broader population.

It is worth noting, however, that across all three datasets (Figures 1, 5, and 6), the highest humor ratings consistently occurred when the symmetry measure was close to zero—indicating that the levels of obviousness and compatibility were nearly equal.

To investigate this further while trying to avoid the issues mentioned above, a subset of puns was manually selected. Each pun was based on an ambiguous expression that was both written and pronounced identically. Additionally, jokes relying on cultural or geographic knowledge were excluded. The result was a dataset of just 22 jokes. Here are two examples:

- *"Why is leather armor best for sneaking? It's literally made of hide"* (score: 1486, symmetry: -0.16)
- *"What do you call a beaten-up pretzel? A salted pretzel"* (score: 5, symmetry: -0.04)

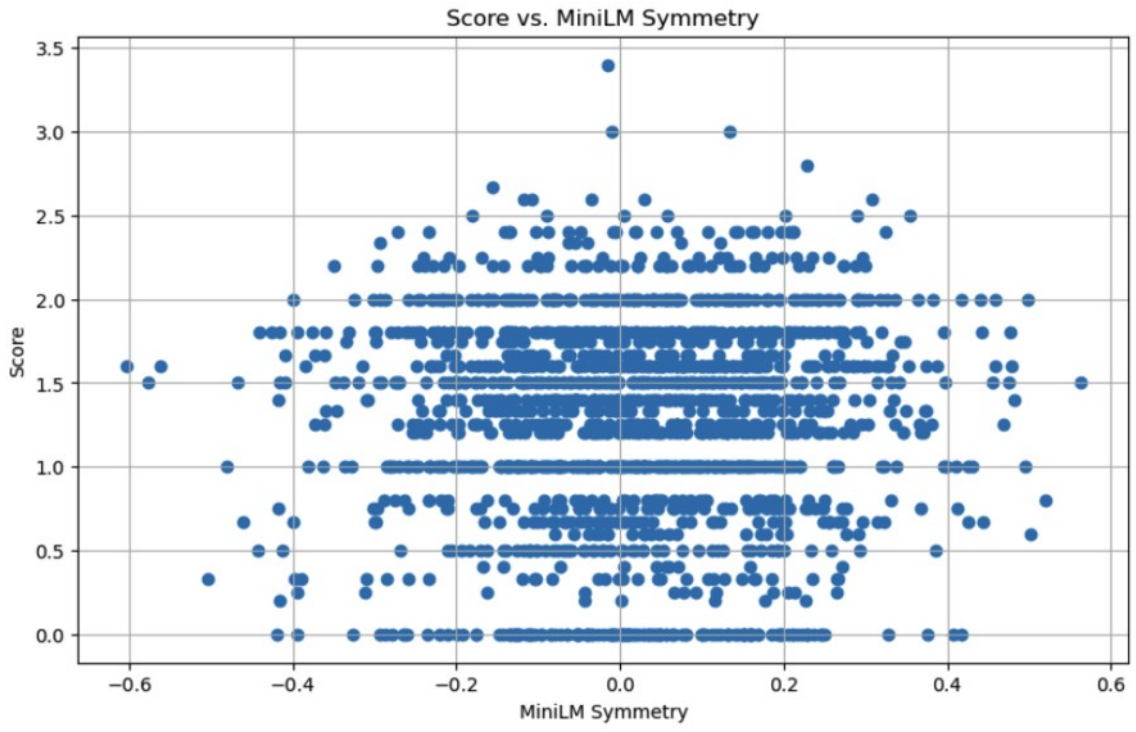

*Figure 5: Score vs. Symmetry (MiniLM model – Expunations dataset)*

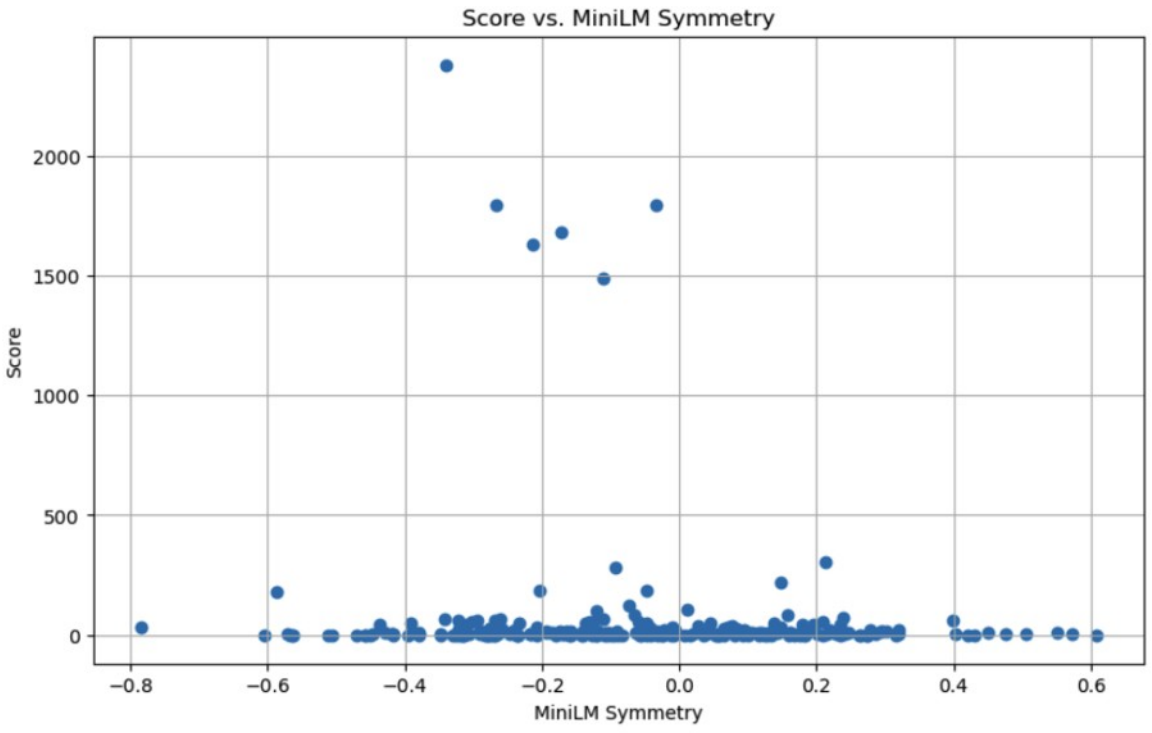

*Figure 6: Score vs. Symmetry (MiniLM model – rJokes dataset)*

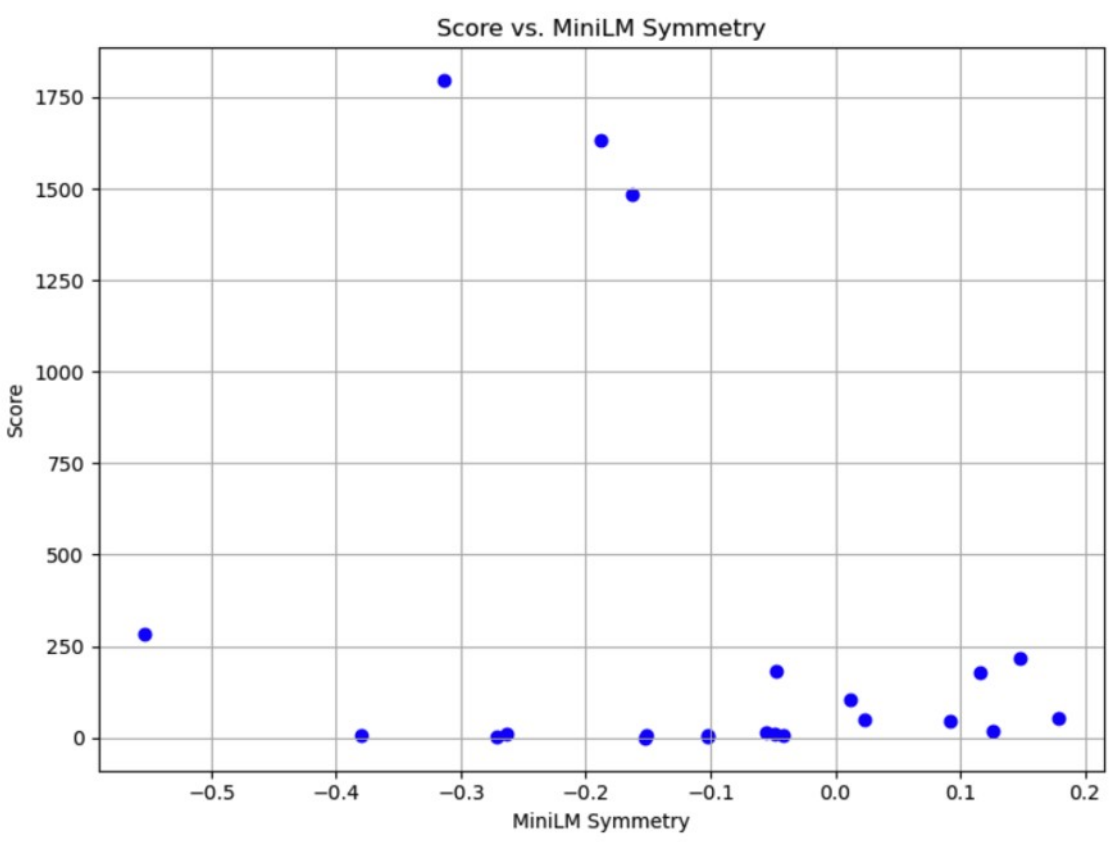

*Figure 7: Score vs. Symmetry*
*(MiniLM model – rJokes dataset - subselection)*

As we see in figure 7, the operation did not significantly increase the quality of the result. The range of symmetry narrowed, but this could simply be a result of the smaller dataset size. However, puns with the highest scores still tend to concentrate around zero symmetry. Across the three datasets, while nearly every joke in the higher score range has a symmetry measure close to zero, those in the lower score range exhibit a wider variety of symmetry values, including some near zero. This suggests that symmetry is a necessary but not sufficient condition for effective humor.

## CONCLUSION

In this paper, four metrics were presented for measuring humor, grounded in incongruity theory. This theory posits that humor arises from a mismatch between expectation—established in the first part of a joke—and surprise, introduced in the second part, which reinterprets earlier elements (e.g., *"How do you stop a lawyer from drowning? Shoot him before he hits the water."*). A variation of the theory suggests that the setup presents a contradiction or surprising premise, and the punchline resolves it in an unexpected way (e.g., *"Why is leather armor best for sneaking? It's literally made of hide."*).

The proposed metrics are obviousness, compatibility, comparison, and symmetry.

- **Obviousness** assesses how intuitive the initial interpretation is.
- **Compatibility** measures how well the punchline aligns with an alternative, hidden meaning.
- **Comparison** ensures the presence of at least two distinct, interpretable readings.
- **Symmetry** captures how evenly both interpretations are expressed, especially when the same word or structure supports both (e.g., *"Coaches usually have a goal in mind"* or *"Old contortionists never die, they just meet their end"*, where "goal" and "meet their end" serve equally well two interpretations of the respective sentences).

These metrics were inspired by Winters *et al.* (2019), who initially proposed five: obviousness, compatibility, comparison, conflict, and inappropriateness—computed using Google Ngrams and Word2Vec. In our revision, conflict was replaced by symmetry and inappropriateness was discarded. The measures originally were taken though Google Ngrams and Word2Vec and we replaced them with word embeddings (calculated through **two models**: OpenAI's *text-embedding-3-small* and MiniLM's *all-MiniLM-L6-v2*).

We took into account **three datasets: JokeJudger, Expunations, and rJokes**. A selection of the last two of them, Expunations and rJokes, was **expanded**. They were processed to select only the jokes that were based on a double meaning expression. For each of the selected jokes, then, two non-humorous sentences were added, representing the two core interpretations of the double meaning expression. For instance, for the joke *"Why did the cook go to jail? He was caught beating the egg,"* the following two sentences were generated: *"She beat the eggs vigorously before adding them to the cake batter"* and *"The suspect was arrested after allegedly beating a man during an argument."* These were used to represent the two senses of "beating." The generated sentences were produced using the model OpenAI *o3-mini-2025-01-31*.

We then calculated the four mentioned metrics for every joke and trained four models—Random Forest, SVM, Naive Bayes, and Linear Regression—to predict humor ratings.

Results showed that the models trained on these metrics performed poorly. On JokeJudger, the baseline accuracy was 61.5%, while our best model achieved only 57.1%. On Expunations and rJokes, where no comparable baseline was available, accuracy dropped further to 51.1% and 41.8%, respectively. Regression analyses revealed no correlation between the metrics and humor ratings.

However, in all datasets, the highest-rated jokes consistently exhibited near-zero symmetry values, indicating that the levels of obviousness and compatibility were closely matched. This suggests that symmetry might be a necessary though not sufficient feature of highly rated jokes.

Several factors likely contributed to the poor predictive performance. A closer look at the datasets suggests that a key issue could be the heterogeneity of the jokes: some rely on wordplay (both including and not including homophones), others on a situational change of perspective, others on stereotypes, references to celebrities, or sexual innuendo—features not uniformly captured by our metrics.

This complexity hints at the need for more advanced approaches. Indeed, Annamoradnejad and Zoghi (2022) demonstrated improved humor prediction by feeding embedding vectors into neural networks with parallel hidden layers, extracting latent relational features. Moreover, large language models like GPT-4o clearly exhibit a capacity for humor understanding and generation. When prompted with *"Tell me a joke about researchers, models, humor and computation,"* GPT-4o responded: *"Why don't computers ever laugh at jokes? Because they always need proper syntax to parse the punchline!"*

When asked to explain, it replied:

> *The joke plays on a pun involving computer terminology:*
>
> - "Syntax"*: In computing, code must follow correct syntax (structure) to be understood or run.*
> - "Parse"*: This is what a computer does to analyze and understand a string (like code or a sentence).*
> - *So the joke imagines that a computer can't laugh at a joke unless the joke has a perfectly structured "syntax" — as if it were code it needs to compile.*

The ability of the model to produce tailored versions of jokes strongly suggests that it learned some mechanisms of humor. Yet, it comes with a crucial limitation: while large models can understand and produce humor, their inner workings do not lend themselves to explanatory insight. In other words, these models can do humor, but cannot yet help humans understand how they do it. This is why even if the use of language models is more efficient, the development of simpler techniques is still very relevant to the field.

## FURTHER WORK

This project can be extended in several directions:

- **Curate a cleaner dataset** by filtering the Expunations dataset to retain only jokes that rely on wordplay, excluding those based on heterographic homophones, stereotypes, celebrity references, or sexual innuendo. The refined narrowed dataset can then be used to reassess the effectiveness of the four metrics in model training, potentially improving predictive performance.
- **Reintroduce a metric for inappropriateness**, to explore whether the presence of taboo or norm-violating content correlates with perceived humor and contributes to rating predictions.
- **Replace machine-generated explanatory sentences** with human-authored sentences, in order to evaluate whether human interpretation of joke meanings improves metric reliability.

Finally, it is noteworthy –and somewhat surprising– that earlier studies using simpler techniques—such as Google Ngrams and WordNet—have achieved better predictive performance than those based on embedding methods. A parallel evaluation of the same jokes using both sets of metrics may offer valuable insights into why this was the case.